\documentclass[a4paper,10pt,twocolumn]{article}

\usepackage[english]{babel}
\usepackage[utf8]{inputenc}
\usepackage[T1]{fontenc}
\usepackage{booktabs}
\usepackage[ruled,vlined]{algorithm2e}

\usepackage[top=1.5cm, left=1.5cm, right=1.5cm, bottom=1.5cm]{geometry}

\renewenvironment{abstract}{\bf\small {\em\ Abstract---}}{}

\usepackage{amsfonts,amssymb,amsmath,amsthm}
\usepackage{subfigure}
\usepackage{graphicx}
\usepackage[footnotesize]{caption}

\usepackage{amsfonts,amssymb,amsmath,amsthm}

\newcommand{\eR}{\mathcal{R}}

\title{Learning Discriminative Representation with Signed Laplacian Restricted Boltzmann Machine}
\author{Dongdong Chen$^1$, Jiancheng Lv$^2$ and Mike E. Davies$^{1}$.\\
  \footnotesize $^1$School of Engineering, University of Edinburgh.\ $^2$College of Computer Science, Sichuan University.} \date{\empty} 

\begin{document}

\maketitle

\begin{abstract}
We investigate the potential of a restricted Boltzmann Machine (RBM) for discriminative representation learning. By imposing the class information preservation constraints on the hidden layer of the RBM, we propose a Signed Laplacian Restricted Boltzmann Machine (SLRBM) for supervised discriminative representation learning. The model utilizes the label information and preserves the global data locality of data points simultaneously. Experimental results on the benchmark data set show the effectiveness of our method.
\end{abstract}

\section{Introduction}\label{sec:introduction}

A restricted Boltzmann Machine (RBM) \cite{smolensky1986information}  is a two layer neural network with one visible layer and one hidden layer. It consists of $m$ visible units $V = (v_1, v_2, \cdots, v_m)$ to represent observable data and $n$ hidden units $H = (h_1, h_2,\cdots,h_n)$ to capture dependencies between observed variables. The RBM has only connections between the layer of hidden and visible variables but not between two variables of the same layer. Accordingly, the random variables $(V, H)$ take values $(v, h)\in \{0,1\}^{m+n}$ and the joint probability distribution under the model is given by the Gibbs distribution
with the energy function:
\begin{equation}\label{eqs:Erbm}
    E(v,h) = -\sum_{i=1}^n\sum_{j=1}^mw_{ij}h_iv_j- \sum_{j=1}^mb_jv_j - \sum_{j=1}^mc_ih_i.
\end{equation}
where $w_{ij}$ is a real valued weight associated with the connection between $v_j$ and $h_i$ and $b_j$ and $c_i$ are real valued bias terms associated with the $j$th visible and the $i$th hidden variable, respectively, $1\leq i\leq n$ and $1\leq j\leq m$. Here we assume that both the visible and hidden units of the RBM are binary, other types of units can be modeled according to \cite{welling2005exponential}.

The RBM has received an increasing amount of interest in recent years because of its excellent ability of unsupervised learning \cite{dongdong2018graph}, and has been successfully adopted in many applications, such as image classification \cite{hinton2006reducing}, document processing \cite{hinton2009replicated}, object segmentation \cite{chen2013deep} and others. More recently, based on the manifold assumption \cite{roweis2000nonlinear,chen2014local}: similar inputs should have a similar representation, a graph regularized RBM (GraphRBM) \cite{dongdong2018graph} was proposed to learn a manifold structure preserved data representation for unsupervised clustering. However, the GraphRBM simply adopt the local neighborhood graph to encoder the locality, so the discriminative ability is limited and the embedding results are not always suitable for the subsequent classification.

\begin{figure}[t]
\begin{center}
    \subfigure[]{\includegraphics[width=0.154\linewidth]{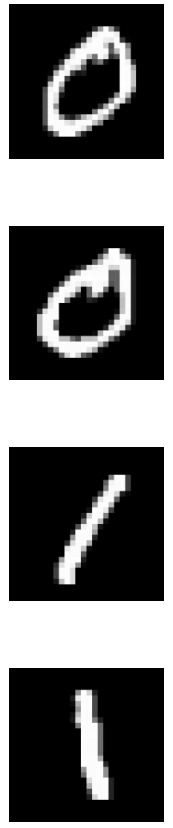}
    \label{fig:1a} }
    \hspace{2ex}
    \subfigure[]{\includegraphics[width=0.75\linewidth]{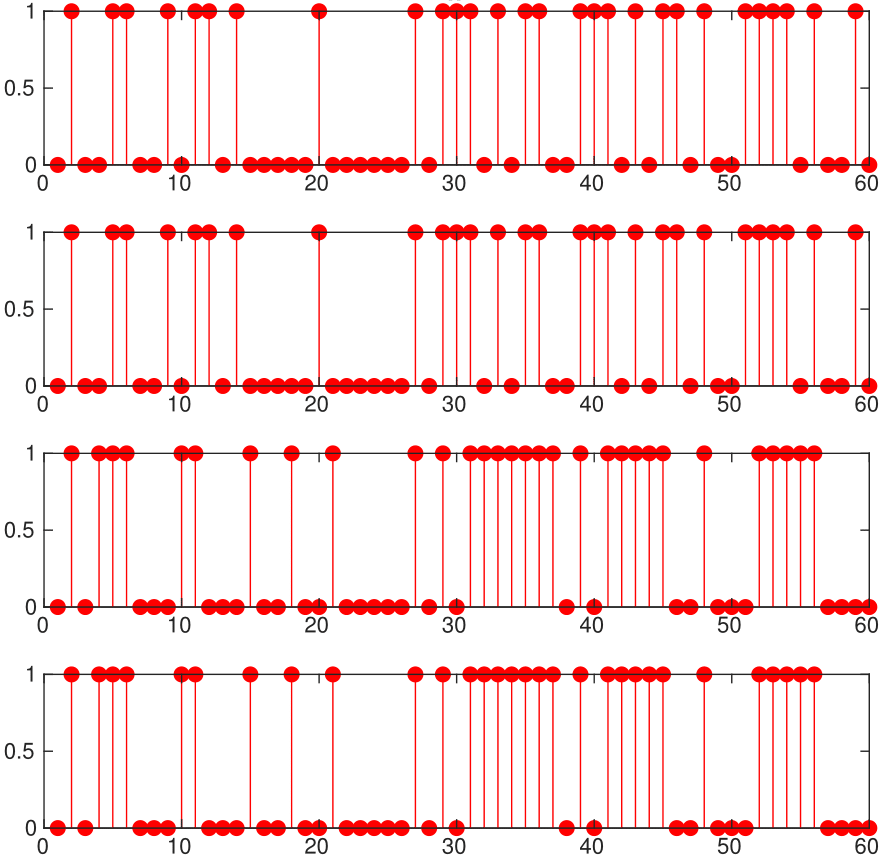}
    \label{fig:1b} }
\end{center}
\caption{Representations learned by SLRBM on the MNIST dataset. (a) Data points from the class '0' and '1'. (b) The corresponding representations.}
\label{fig:1}
\end{figure}

In this abstract, we extend the GraphRBM to investigate the potentials of RBM for learning  discriminative representations. By constructing the binary signed graph and employing the signed graph Laplacian, we train a new RBM based model, dubbed as the Signed Laplacian Restricted Boltzmann Machine (SLRBM) for learning discriminative representation (Figure \ref{fig:1}). The SLRBM utilizes the label information and preserves the global data locality of data points simultaneously.
We show that 1) the data points belonging to the same class, not simply originally nearby, are better projected together for the subsequent classification;
2) Compared with the GraphRBM, the elements in the adjacency matrix are allowed to take negative values to enable the incorporation of both similarity and dissimilarity information, so that better discriminative performance can be achieved; 3)
Different from the traditional manifold learning methods \cite{roweis2000nonlinear,chen2017unsupervised,chen2017angle}, the nearest neighborhood search is not required which makes the SLRBM easier to implement.

\section{Proposed Method}

\subsection{Signed Graph Laplacian}
The local geometric structure of data can typically be modeled through a nearest neighbor graph \cite{roweis2000nonlinear,chen2014local,zhu2018ldmnet}. Consider a graph with $N$ vertices to describe the geometric structure of data, where each vertex corresponds to a data point. For each data point $x_i$, we find its $p$ nearest neighbors and put edges between $x_i$ and its neighbors. There are many choices to define the adjacency matrix $\Phi=[\phi_{ij}] (i, j = 1,2,\cdots, N)$ on the graph \cite{yan2007graph}. For example, $\phi_{ij}$ can be defined by the Gaussian heat kernel distance, i.e.,  if nodes $i$ and $j$ are connected, the graph weight is computed as  $\phi_{ij}=\exp(-\|x^{(i)}-x^{(j)}\|^2/\rho)$, where $\rho$ is the kernel width. Binary weight is another commonly used strategy in which $\phi_{ij} = 1$ if and only if nodes $i$ and $j$ are connected by an edge, $\phi_{ij} = 0$, otherwise.

To introduce more discriminative information into the graph construction, a signed graph was recently proposed in \cite{kunegis2010spectral} which defines a novel adjacency matrix $\Phi$ containing both positive and negative elements. Specifically, the vertices in the graph correspond to the training data points $\{x^{(i)}\}_{i=1}^N$. $x^{(i)}$ and $x^{(j)}$ are connected by a positive edge if they belong to the same class, while they are linked by a negative edge if they come from different classes. Therefore, the elements in the adjacency matrix $\Phi$ are
\begin{equation}
\phi_{ij}=\left\{
\begin{array}{rl}
1, & label(x^{(i)}) = label(x^{(j)});
\\
\\
-1, &label(x^{(i)}) \neq label(x^{(j)});.
\end{array}
\right.
\end{equation}

By defining a diagonal matrix $D$ as $D_{ii} = \sum_j\phi_{ij}$ for $i=1,2,\cdots,n$, the \emph{signed graph Laplacian} is formulated as accordingly
\begin{equation}
L = D - \Phi,
\end{equation}
it is easy to verify that $L$ is positive semi-definite.

\subsection{SLRBM}
Given a data set $\mathcal{X} = \left(x^{(1)}, x^{(2)},\cdots,x^{(N)}\right)\in \eR^{m\times N}$, its corresponding hidden representations learned by RBM is denoted by $\mathcal{H} = (h^{(1)},h^{(2)},\cdots,h^{(N)})\in\eR^{n \times N}$. According to \cite{dongdong2018graph}, the performance of structure preservation in the hidden representation space (i.e. the smoothness of the hidden representation) can be measured by the following term:
\begin{equation}\label{eqs:E_graphMatrix}
\mathcal{J}(\mathcal{H}) = \sum\limits_{ij}\phi_{ij}\|h^{(i)} - h^{(j)}\|^2.
\end{equation}
where the pairwise Euclidean distance $\|h^{(i)} - h^{(j)}\|^2$ is used to measure the dissimilarity between $h^{(i)}$ and $h^{(j)}$, $i,j=1,2,\cdots,N$.

By minimizing (\ref{eqs:E_graphMatrix}), we expect that if two data points $x^{(i)}$ and $x^{(j)}$ come from the same class (i.e. $\phi_{ij}=1$ ), $h^{(i)}$ and $h^{(j)}$ are also close to each other, and vice versa. By incorporating the signed Laplacian regularizer (\ref{eqs:E_graphMatrix}) with the original RBM energy function \ref{eqs:Erbm}, we can get the energy function of SLRBM. Specifically, the sample-wise energy function of our model is defined as follows:

\begin{equation}\label{eqs:E_GraphRBMfull}
F(v^{(i)},h^{(i)}) = E(v^{(i)},h^{(i)}) + \lambda\sum_{j}\phi_{ij}\|h^{(i)} - h^{(j)}\|^2.
\end{equation}
where $\lambda \geq 0$ is the regularization trade-off parameter which controls the smoothness of hidden representations.

Define $\delta^{(i)}=\sum_{j}\phi_{ij}(h^{(i)}-2h^{(j)})$. Then, the conditional probabilities for each visible and hidden unit given the others are

\begin{align}
p(h_{l}^{(i)}=1|v^{(i)})  =& f ( \sum\limits_{r} W_{lr}v_{r}^{(i)} + c_l -  \delta_{l}^{(i)});\label{eqs:P_HV_graphRBM}\\
 p(v_{l}^{(i)}=1|h^{(i)}) =& f ( \sum\limits_{s} W_{sl}h_{s}^{(i)} + b_l ),\label{eqs:P_VH_graphRBM}
\end{align}
where $\delta_{l}^{(i)} = \lambda \sum_{j}\phi_{ij}(1-2h_{l}^{(j)})$.

The gradient of $\theta=\{W, b, c\}$ can  be computed using the contrastive divergence (CD) \cite{hinton2002training} algorithm accordingly. Finally, the network parameter $\theta$ is updated using the standard gradient ascent strategy.

\subsection{Relation to Other Methods}
It is easy to verify that the signed graph Laplacian regularizer  $\mathcal{J}(\mathcal{H})=\sum_{ij}\phi_{ij}\|h^{(i)} - h^{(j)}\|^2 = 2\emph{\emph{Trace}}(\mathcal{H}L\mathcal{H}^{\top})$, which implies that minimizing (\ref{eqs:E_graphMatrix}) is a standard graph embedding problem formulated in  \cite{yan2007graph}. Therefore, the regularizer $\mathcal{J}(\mathcal{H})$ essentially provides a manifold separation over the graph.

According to the patch alignment framework proposed in \cite{zhang2009patch}, various dimension reduction algorithms, such as \cite{roweis2000nonlinear,jolliffe1986principal}, can be summarized into a one unified formulation. Following the tricks in \cite{kunegis2010spectral,zhang2009patch}, and let $\phi^{(i)}$ denote the $i$th column of $\Phi$, it is easy to verify that the signed Laplacian can be regarded as a special case of patch alignment framework if
\begin{equation}
    L = \left(
  \begin{array}{cc}
    D_{ii} & -(\phi^{(i)})^{\top}\\
    -\phi^{(i)}& \Gamma^{(i)}\\
  \end{array}
\right),
\end{equation}
where $\Gamma^{(i)} = diag(|\phi^{(i)}_1|, \cdots, |\phi^{(i)}_N|)$ is a diagnal matrix.

Specifically, the patches of  SLRBM are globally constructed by using all the data points in the dataset, while the GraphRBM establishes each patch by a  data point and its nearest neighbors. SLRBM preserves the proximity relationship in a patch through the adjacency matrices, which are different from GraphRBM that preserves local coefficients obtained in the original high-dimensional space.

\section{Experiment and Discussion}
We evaluate the performance of the SLRBM using the benchmark MNIST dataset \cite{lecun1998gradient} for the task of discriminative representation learning.\footnote{Experiment setting are as follows: batch size 100, weight decay $10^{-4}$, 100 epochs.} This MNIST includes 60,000 handwritten digits samples used for training and 10,000 samples for testing, with an image size of $28\times28$ binarized grayscale pixels.

\begin{table}[h]
 \centering
  \fontsize{8.4}{10.5}\selectfont
\begin{tabular}{l|l }
Model          & Error  \\\hline
SLRBM ($\lambda = 10^{-2}$, $\eta = 10^{-2}$)     & 11.2\%\\\hline
ClassRBM & 23.8\%\\
RBM & 26.8\%\\
\end{tabular}
  \caption{Classification performances for the different models, $\eta$: learning rate.}\label{tab:results}
\end{table}

Figure \ref{fig:1} illustrates the representations learnt by SLRBM. It shows the representations of two different digit '0' (or '1') are almost the same, while the representations of digits '0' and '1' are totally different. We are further interested in knowing whether the representations learnt by SLRBM are useful for representation based image classification \cite{vidal2011subspace}. Consequently, the nearest neighbor classifier (1NN) is applied on the learned hidden representations to classify the testing data and compute the error rates. Table \ref{tab:results} shows the classification results of two counterpart models namely RBM, ClassRBM \cite{larochelle2012learning} and our SLRBM. Note that we did not conduct the additional supervised fine tuning. The result justifies the importance of considering the signed Laplacian regularizer on RBM for learning discriminative data representation, which demonstrates the motivation of this study: the data points belonging to the same class, not simply originally nearby, are better to be projected together for the subsequent classification; the elements in the adjacency matrix are allowed to take negative values to enable the incorporation of both similarity and dissimilarity information, so that better discriminative performance can be achieved.

\section*{Acknowledgment}
This work is partially funded by the ERC C-SENSE project (ERCADG2015-694888). Jiancheng Lv is funded by National Nature Science Foundation of China (Grant No. 61375065 and 61625204).

\bibliographystyle{IEEEtran}
\bibliography{itwist_rbm}

\end{document}